\documentclass[runningheads]{llncs}

\usepackage{eccv}

\usepackage{eccvabbrv}

\usepackage{graphicx}
\usepackage{booktabs}

\usepackage{graphicx}
\usepackage{subfloat}
\usepackage{multirow}
\usepackage{algorithm}
\usepackage{algorithmic}
\usepackage{booktabs}
\usepackage{stfloats}
\usepackage{caption}
\usepackage{epstopdf}
\usepackage{color}
\usepackage{colortbl}
\usepackage{rotating}
\usepackage{mathrsfs}
\usepackage{makecell}
\usepackage{wrapfig}
\usepackage{marvosym}

\usepackage[accsupp]{axessibility}  

\usepackage{hyperref}

\usepackage{orcidlink}

\begin{document}

\title{VCP-DCN: Beyond Visual Concealed Property via Depth Collaborative Network for Camouflaged Object Detection} 

\titlerunning{VCP-DCN}

\author{Songsong Duan\orcidlink{0000-0003-2983-4044} \and
Xi Yang\textsuperscript{\Letter}\orcidlink{0000-0002-5791-3674} \and
Nannan Wang\orcidlink{0000-0002-4695-6134}
}

\authorrunning{Duan et al.}

\institute{the State Key Laboratory of Integrated Services Networks, \\ School of Telecommunications Engineering, \\ Xidian University, Xi'an, China \\
\email{duanss@stu.xidian.edu.cn, yangx@xidian.edu.cn, nnwang@xidian.edu.cn}
}

\maketitle

\begingroup
\renewcommand{\thefootnote}{}
\footnotetext{\textsuperscript{\Letter}Corresponding author: Xi Yang.}
\addtocounter{footnote}{-1}
\endgroup

\begin{abstract}
  Camouflaged Object Detection (COD) aims to identify and segment camouflaged objects in complex environments, which are often concealed because their color and texture are similar to the background. Several existing COD methods introduce depth maps to boost detection performance via learning complementary RGB-D features, ignoring modality-specific characteristics of concealed objects in the depth domain. To address this issue, we propose a depth collaborative network, called VCP-DCN, to mine distinguishable multi-modality features beyond visual concealed prototype in depth domain. Specifically, VCP-DCN progressively performs multi-modality alignment, interaction, and fusion for the COD task. In the \textbf{alignment} stage, we propose a Separable Prototype Embedding (SPE) module to learn modality-consistency and modality-specific RGB/depth prototype tokens through prototype contrastive learning. Furthermore, we develop a Multi-modality Dual Attention (MDA) module to enhance the cross-modal feature representation through local response maps between modality-consistency RGB/depth prototype tokens and visual tokens on the \textbf{interaction} stage. Finally, we design a Depth Adaptive Injection (DAI) module to adaptively measure contribution of RGB/depth features with a decision-making mechanism, which calculates similarity distance between RGB/depth modality-specific prototype tokens and modality-consistency ones on the \textbf{fusion} stage. Extensive experiments demonstrate the effectiveness of our VCP-DCN on three authoritative datasets. The code is available at \href{https://github.com/duan-song/VCPDCN}{https://gith\\ub.com/duan-song/VCPDCN}.
  
\end{abstract}

\section{Introduction}
\label{sec:intro}

In nature, some camouflaged animals, such as chameleons and geckos, possess innate defense mechanisms that help them evade predators. From a biological perspective, these animals adapt their coloration, morphology, and other attributes to blend into their surrounding environment for concealment purposes \cite{ref-1}. Drawing inspiration from these natural strategies, camouflaged object detection (COD) studies the concealment property in real-world scenes by searching and segmenting camouflaged objects. As a result, COD has become a vital research field in computer vision, with a wide range of applications including underwater understanding \cite{ref-2}, agricultural crop detection \cite{ref-3}, and medical image segmentation \cite{ref-4}.

\begin{figure}[t]
	\centering
	\includegraphics[width=0.980\textwidth,height=3.8cm]{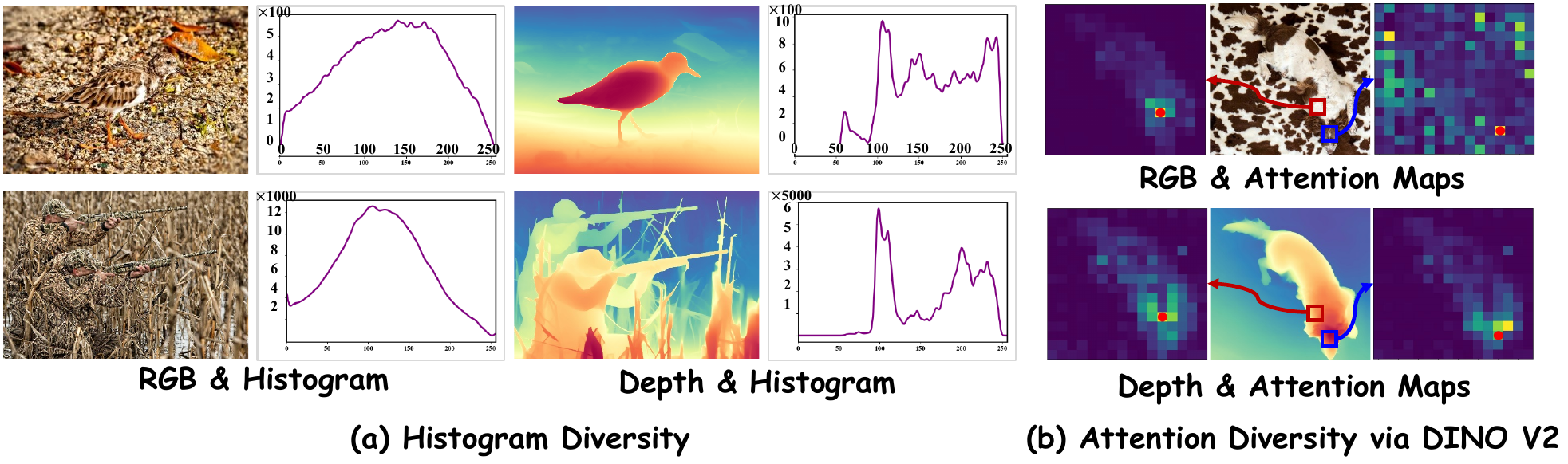}
	\caption{Comparison with RGB and Depth images via histogram and attention maps.}
	\label{Fig.1}	
\end{figure}

\begin{figure}[t]
	\centering
	\includegraphics[width=0.980\textwidth,height=3.8cm]{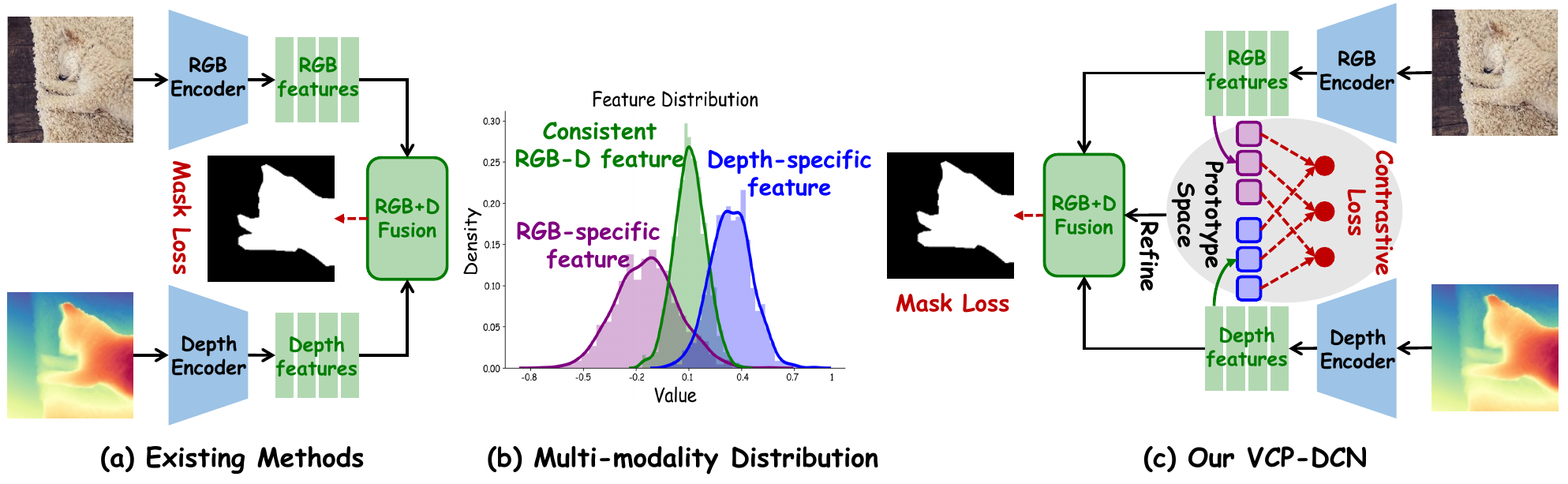}
	\caption{Comparison with existing RGB-D COD and our VCP-DCN. (a) is the existing RGB-D COD methods, which learns the RGB-D complementary features under the supervision of mask loss; (b) is the illustration of multi-modality feature distribution; (c) is our VCP-DCN, which learns multi-modality specific features to refine the RGB-D features via separable prototype learning.}
	\label{Fig.1}	
\end{figure}

However, in COD, the low discrepancy between camouflaged objects and the background, \textit{e.g.}, scale, color, appearance, and occlusion, poses significant challenges from a visual-domain perspective, often leading to unreliable accuracy. Our analysis reveals that existing COD methods struggle to produce fine-grained localization masks for camouflaged objects due to their inherent visual concealment. As illustrated in Fig. 1 (a), we analyze the pixel intensity distribution of camouflaged images using histograms. These histograms exhibit a prominent single peak, indicating a lack of distinct brightness stratification and characterizing the images as low-contrast. Beyond image-level analysis, we also investigate representation learning to further demonstrate the visual concealment property. Fig. 1 (b) visualizes attention maps derived from DINOv2 \cite{ref-11} seed tokens applied to camouflaged images. The results show that attention maps originating from both foreground and background seeds emerge ambiguous localization information.

Given that camouflaged objects exhibit concealed properties in the visual domain, \textit{\textbf{can we explore novel characteristics of camouflaged objects from alternative perspectives?}} To this end, we study the feasibility of depth maps for the COD task. We compare the visual cues, histogram, and attention maps of depth maps with camouflaged images in Fig. 1. The comparison indicates that depth maps provide salient spatial and geometry priors, which is an effective supplement for the COD task. Based on this analysis, some works work \cite{ref-64, ref-65, ref-3, ref-38, ref-50} have also attempted to introduce depth maps into COD community. However, the existing methods are limited by the \textbf{feature optimization bias} caused by the multi-modality fusion stage, where RGB and Depth features tend to represent RGB-D consistent features. As shown in Fig. 2 (a) and (b), the unified supervision of GT mask leads to the convergence of multi-modality feature representations and the loss of inherent modal discriminability. Eventually, the RGB features and Depth features become homogeneous representations, which fail to exert their complementary modal advantages. \textbf{\textit{Therefore, how to effectively integrate depth information for in-depth exploration of the COD task remains challenging.}}

To address the feature optimization bias, this paper proposes a depth collaborative network, termed VCP-DCN, to learn enrich diversity features in visual and depth domains, where we project RGB-D features into modality-aware prototype tokens to reserve modal discriminability in multi-modality fusion stage under supervision of contrastive loss in Fig. 2 (c). Specifically, VCP-DCN develops a progressive multi-modality presentation learning strategy with multi-modality alignment, interaction, and fusion. In the alignment stage, we propose a Separable Prototype Embedding (SPE) module to extract multi-modality foreground and background prototype tokens, which are further projected into modality-consistency and modality-specific prototypes. We employ prototype contrastive loss to optimize prototype representation of the multi-modality features. In the interaction stage, we propose a Multi-modality Dual Attention (MDA) module to enhance RGB and depth features through exchanging cross-modal prototype semantic maps generated by global-to-local matching between modality-consistency prototypes and visual features. MDA further employs a masked linear attention to refine foreground and background multi-modality features. In the fusion stage, we propose a Depth Adaptive Injection (DAI) module to dynamically fuse RGB and Depth features by measuring the similarity distance between modality-specific and modality-consistency prototypes. Extended experiments on three benchmark datasets demonstrate that our VCP-DCN outperforms previous state-of-the-art COD methods. Our main contributions are summarized as:
	
(1) We propose a novel perspective to explore camouflaged object detection from multi-modality feature optimization bias, and propose a depth collaborative network, termed VCP-DCN, to mine salient cues beyond visual concealed property.

(2) We propose a progressive multi-modal leaning strategy, which consists of three key parts: SPE module for multi-modal alignment, MDA for multi-modal interaction, and DAI module for multi-modal fusion. 

(3) We innovatively introduce prototype contrastive loss to the COD community, which is utilized to reserve modality-aware discriminative cues and avoid RGB-D homogenization representation in  multi-modality feature fusion stage.

\section{Related Work}
\label{sec:formatting}

\subsection{Camouflaged Object Detection}
COD is a challenging vision task for understanding camouflaged scenes, which aims to search, recognize, and segment camouflaged objects. Fan \textit{et al.} \cite{ref-14} first systematically introduced camouflage object detection and proposed a new large-scale COD10K dataset. Further, SINet \cite{ref-15} proposed a biological-inspired detection framework via search and localization manners. After the pioneer works \cite{ref-14, ref-15}, COD has been attracting growing attention. 

The development of COD \cite{ref-16, ref-17} is closely related to the progress of neural networks. In the early stage, researchers usually used ConvNet \cite{ref-5, ref-7, ref-8} to extract the multi-scale local features. Zhou \textit{et al.} \cite{ref-18} used multi-scale convolution operations to extract boundary characteristics for capturing the scale variations of the camouflaged objects. Unlike ConvNet, vision transformer \cite{ref-6, ref-9, ref-22} has attracted much attention due to its superior long-range modeling ability and has been widely applied in the COD community. Huang \textit{et al.} \cite{ref-19} proposed a transformer-based feature shrinking pyramid network to hierarchically learn locality-enhanced neighboring transformer features. Different from the pyramid network, Cong \textit{et al.} \cite{ref-20} proposed a frequency perception network to model semantic hierarchy in the frequency domain. Furthermore, Luo \textit{et al.} \cite{ref-21} introduced multi-task learning into the COD community, which innovatively designed a general 2D visual prompt learning to unify the salient and object detection tasks. Besides the pre-trained transformer, some large vision foundation models were progressively employed in COD community. Zhao \textit{et al.} \cite{ref-23} introduced the stable diffusion model \cite{ref-24} into the COD community and proposed a FocusDiffuser to study how generative models can improve camouflaged object detection.

Unfortunately, the computational complexity of the vision transformer severely restricts the application of the COD methods in real-time scenarios. To address this issue, some works introduced state space model (SSM) \cite{ref-25, ref-26, ref-27}, a new generation of neural networks, into the COD community. He \textit{et al.} \cite{ref-28} proposed a unified saliency detection framework with any modality, including RGB, RGB-Depth, RGB-Thermal, and Video data. Zhang \textit{et al.} \cite{ref-29} utilized causal inference to mine part-whole relational property with a Mamba model. 

%

\subsection{Camouflaged Object Detection with Depth}
Depth maps are usually viewed as vital and effective auxiliary information for challenging vision tasks, including RGB-D semantic segmentation \cite{ref-30, ref-31}, RGB-D salient object detection \cite{ref-33, ref-32}, and RGB-D object tracking \cite{ref-34, ref-35}. Because depth maps contain clear boundary cues and spatial layout priors, some works attempt to utilize depth to assist COD tasks. Liu \textit{et al.} \cite{ref-36} proposed a dual stream adapter within Segment Anything Model \cite{ref-37}, which used a parallel manner to control RGB and depth adapters for the COD task. Yu \textit{et al.} \cite{ref-38} designed a prompt deeper module to interact the RGB and depth features with bias correction. Wu \textit{et al.} \cite{ref-39} explored the objects' ``pop-out'' prior in depth maps, which assumed objects reside on the background surface. Wang \textit{et al.} \cite{ref-40} proposed a depth-aided COD method with a cross-modal asymmetric fusion module. In addition to the COD community, depth maps was widely used in many binary mask segmentation, like salient object detection \cite{ref-41, ref-42, ref-43}, polyp segmentation \cite{ref-44, ref-45}, and anomaly detection \cite{ref-46, ref-47}. 

Although early works attempted to introduce depth maps into the COD community, their accuracy was usually not as good as that of the RGB COD methods. We believe that this gap is caused by two factors: inconsistent multimodal cues and an inefficient fusion strategy. To address this problem, we first utilized a depth foundation model (Depth Anything \cite{ref-13}) to generate consistent and high-quality depth maps. Then, we proposed a progressive multi-modal learning strategy to align, interact, and fuse multi-modal features.

\section{Methodology}

\subsection{Overall Architecture}
Following common practical network architecture \cite{ref-22, ref-48, ref-55}, we adopt an end-to-end encoder-decoder structure to build our VCP-DCN. As shown in Fig. 3, our VCP-DCN consists of an encoder, a decoder, and the proposed depth collaborative networks.

\textbf{Encoder}: To extract RGB and Depth features, we introduce a VMamba-S \cite{ref-25} as the encoder. The RGB and Depth encoders share parameters. Given an input image $\mathcal{I}_r \in \mathbb{R}^{3 \times H \times W}$ and corresponding depth map $\mathcal{I}_d \in \mathbb{R}^{1 \times H \times W}$, we feed it into the shared encoder to generate hierarchy pyramid features, which denote as $f_r = \{f_{r}^{i}\}_{i=1}^{4}$ and $f_d = \{f_{d}^{i}\}_{i=1}^{4}$. Note that the spatial resolutions of hierarchy pyramid features are $\mathbb{R}^{C_{i} \times H/2^{i+1} \times W/2^{i+1}}$, where the channel dimensions are $C_{i} \in \{96, 192, 384, 768\}$. Next, we embed the proposed depth collaborative network into each layer of hierarchy pyramid features to generate modality features $\mathcal{F}_f = \{\mathcal{F}_{f}^{i}\}_{i=1}^{4}$ with the same spatial resolutions and channel dimensions.

\textbf{Decoder}: We follow a triple decoder structure to generate three prediction masks from RGB, Depth, and RGB-D decoders, which has been proven to be an effective design in previous works \cite{ref-58, ref-59} for RGB-D segmentation tasks. Specifically, the multi-modality pyramid features $f_r$, $f_d$, and $\mathcal{F}_f$ are feed into the triple decoders to obtains prediction masks $\mathcal{M}_r$, $\mathcal{M}_d$, and $\mathcal{M}_f$. To achieve an effective decoder, we use 1$\times$1 convolution and 2$\times$ up-sample operations to connect cross-level features. 

\begin{figure*}[t]
	\centering
	\includegraphics[width=0.90\textwidth,height=8cm]{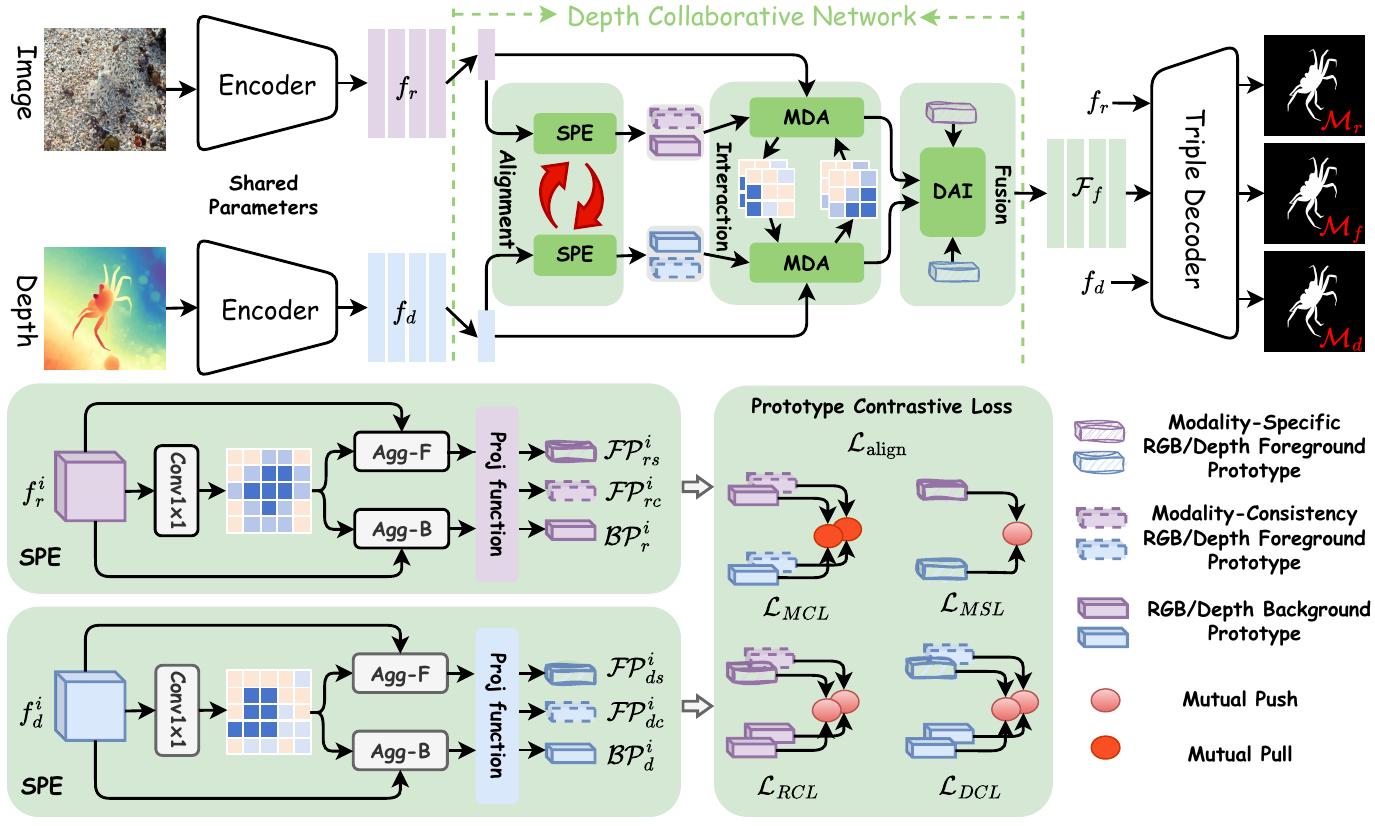}
	\caption{Illustration of the proposed VCP-DCN, which consists of Separable Prototype Embedding (SPE), Multi-modality Dual Attention (MDA), Depth Adaptive Injection (DAI) modules. In the alignment stage, a prototype contrastive loss ($\mathcal{L}_{\mathrm{align}}$) is proposed to lean modality-consistency and modality-specific prototypes, which contains $\mathcal{L}_{MCL}$, $\mathcal{L}_{MSL}$, and $\mathcal{L}_{RCL}$.}
	\label{Fig.2}	
\end{figure*}


\subsection{Separable Prototype Embedding}
Our VCP-DCN employs a progressive strategy to learn multi-modality feature presentation with multi-modality alignment, interaction, and fusion. As shown in Fig. 3, we propose an SPE module to achieve multi-modality alignment. For the RGB-D segmentation task, a conventional operation is used to indiscriminately fuse RGB and Depth features for the subsequent decoder. However, this indiscriminate fusion strategy ignores diversity between RGB and Depth features from the perspective of global optimization. In COD community, several methods \cite{ref-64, ref-65, ref-3} adopt indiscriminate fusion approach to extract complementary RGB-D features, ignore the diversity of RGB and Depth features.

To address this issue, the SPE module projects RGB and Depth features into global prototype tokens via decoupling foreground and background features. The decoupled multi-modality foreground and background prototypes are further translated into the modality-consistency and modality-specific prototypes for representing diversity between RGB and Depth features:
\begin{equation}
	\begin{split}
		\mathcal{SM}_r^i &= \mathrm{Sigmoid}(\mathrm{Proj}_{\mathcal{C}}(f_r^i)), \\
		\mathcal{FG}_r^i &=\frac{1}{H_i \times W_i} \sum_{m=1}^{H_i}\sum_{n=1}^{W_i} \mathrm{Mul}(\mathcal{SM}_r^i, f_r^i)_{m,n} , \\
		\mathcal{BG}_r^i & =\frac{1}{H_i  \times W_i} \sum_{m=1}^{H_i}\sum_{n=1}^{W_i} \mathrm{Mul}(1-\mathcal{SM}_r^i, f_r^i)_{m,n} ,
	\end{split}
\end{equation}
where $\mathrm{Proj}_{\mathcal{C}}$ is a 1$\times$1 convolution, which is leveraged to generate foreground semantic map $\mathcal{SM}_r^i \in \mathbb{R}^{1\times H_i \times W_i}$ via compressing feature channels. $H_i = \frac{H}{2^{i+1}}$ and $W_i = \frac{W}{2^{i+1}}$ are height and width of $f_r^i$. $\mathrm{Sigmoid}$ is activation function of $\mathcal{SM}_r^i$. $\mathcal{FG}_r^i$ and $\mathcal{BG}_r^i$ are global foreground and background tokens of RGB features. Furthermore, we use MLP operations to learn modality-consistency and modality-specific prototype tokens:
\begin{equation}
	\begin{split}
		\mathcal{FP}_{rs}^i &= \mathrm{MLP}_{rs}(\mathcal{FG}_r^i), \\
		\mathcal{FP}_{rc}^i &= \mathrm{MLP}_{rc}(\mathcal{FG}_r^i), \\
		\mathcal{BP}_{r}^i &= \mathrm{MLP}_{rb}(\mathcal{BG}_r^i),
	\end{split}
\end{equation}
where $\mathrm{MLP}_{rs}$, $\mathrm{MLP}_{rc}$, and $\mathrm{MLP}_{rb}$ represent three MLP operations with different projection functions for modality-specific foreground prototype $\mathcal{FP}_{rs}^i$, modality-consistency foreground prototypes $\mathcal{FP}_{rc}^i$, and modality-consistency background prototypes $\mathcal{BP}_{r}^i$. For the depth branch, we use the same processing to extract $\mathcal{FP}_{ds}^i$, $\mathcal{FP}_{dc}^i$, and $\mathcal{BP}_{d}^i$ from the depth feature $f_d^i$ through Eq. (1) and (2).

To learn modality-specific and modality-consistency presentation, we introduce contrastive learning to model and optimize the multi-modality prototype space. Specifically, we narrow the similarity distance between $\mathcal{FP}_{rc}^i$ and $\mathcal{FP}_{dc}^i$ for modality-consistency foreground prototype tokens. Similarly, we implement the same processing on background features to learn modality-consistency background prototype tokens:
\begin{equation}
	\begin{split}
		\mathcal{L}_{MCL} &=- \sum_{i=1}^{4} log \,\, ( \mathrm{Cos}(\mathcal{FP}_{rc}^i, \mathcal{FP}_{dc}^i) \\
		& \quad \,\, + \mathrm{Cos}(\mathcal{BP}_{r}^i, \mathcal{BP}_{d}^i)),
	\end{split}
\end{equation}
where $\mathrm{Cos}(,)$ is the cosine similarity of two prototype tokens. $\mathcal{L}_{MCL}$ is the modality-consistency contrastive loss for RGB and Depth features. After $\mathcal{L}_{MCL}$, we use modality-specific contrastive loss $\mathcal{L}_{MSL}$  to enlarge the similarity distance between $\mathcal{FP}_{rs}^i$ and $\mathcal{FP}_{ds}^i$ for modeling the modality-specific prototypes:
\begin{equation}
	\mathcal{L}_{MSL} =- \sum_{i=1}^{4} log \,\, (1-\mathrm{Cos}(\mathcal{FP}_{rs}^i, \mathcal{FP}_{ds}^i)).
\end{equation}

Under the constraints of  $\mathcal{L}_{MCL}$ and $\mathcal{L}_{MSL}$, we learn modality-consistency and modality-specific representation by  optimizing the similarity distance between multi-modality prototypes. Besides, we discover that foreground prototypes and background prototypes are opposed to each other in the feature space via the contrastive losses  $\mathcal{L}_{RCL}$ and  $\mathcal{L}_{DCL}$ between foreground and background prototype tokens:
\begin{equation}
	\begin{split}
		\mathcal{L}_{RCL} &=- \sum_{i=1}^{4} (log \,\, (1-\mathrm{Cos}(\mathcal{FP}_{rs}^i, \mathcal{BP}_{r}^i)) \\
		& \quad \,\, + log \,\,(1-\mathrm{Cos}(\mathcal{FP}_{rc}^i, \mathcal{BP}_{r}^i))).
	\end{split}
\end{equation}

By replacing the RGB prototypes in Eq. (5) with Depth prototypes, we can obtain the contrast loss $\mathcal{L}_{DCL}$ for Depth features.

\begin{figure*}[t]
	\centering
	\includegraphics[width=0.980\textwidth,height=6cm]{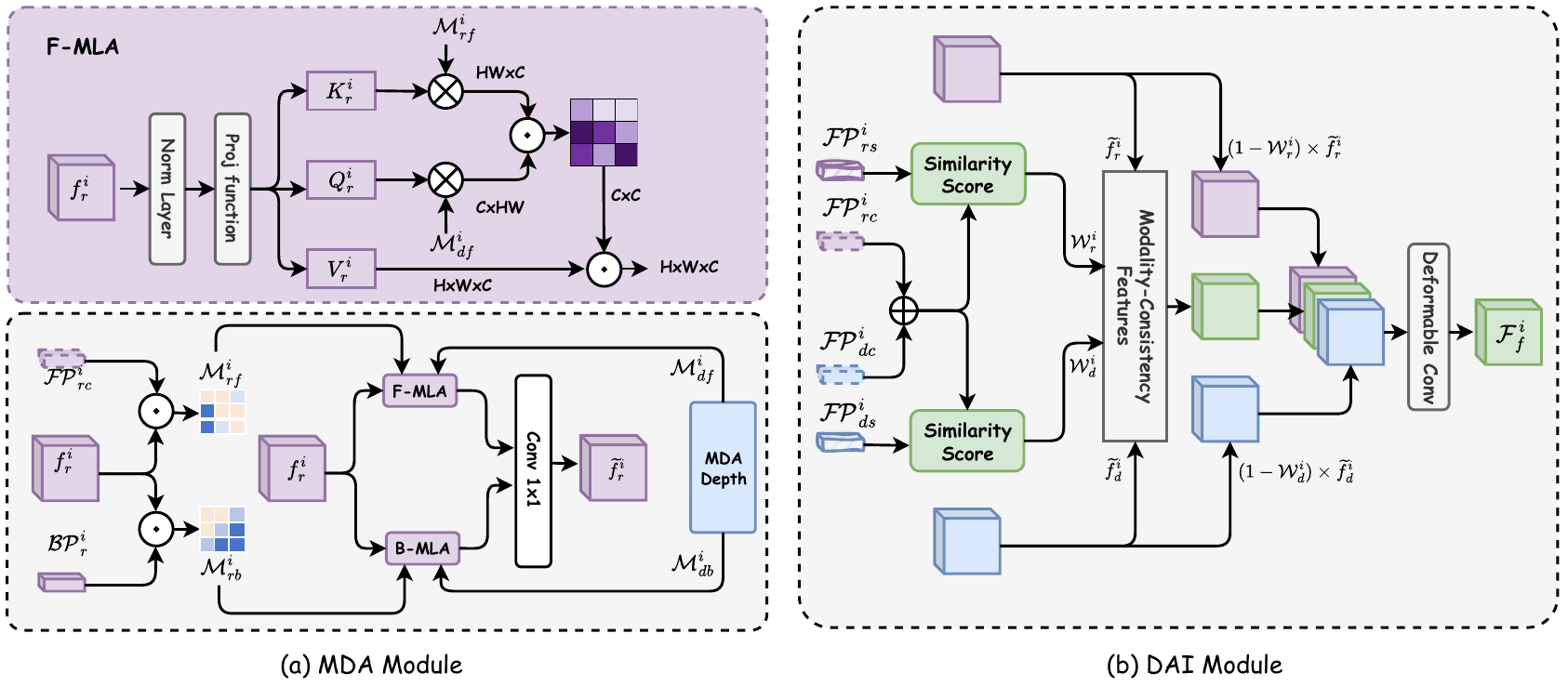}
	\caption{Illustration of MDA module of RGB feature and DAI module. Depth feature shares a similar MDA structure.}
	\label{Fig.3}	
\end{figure*}

\subsection{Multi-modality Dual Attention}
After the SPE module, we obtain modality-consistency RGB and Depth prototypes, which are further utilized to achieve interaction between RGB and Depth features in the MDA module. As shown in Fig. 4 (a), the MDA module first generates foreground and background masks through global-to-local token matching between global prototypes and RGB/Depth features. To illustrate this, consider the RGB branch:
\begin{equation}
	\mathcal{M}_{rf}^i = f_r^i \odot \mathcal{FP}_{rc}^i,  \quad
	\mathcal{M}_{rb}^i = f_r^i \odot \mathcal{BP}_{r}^i,
\end{equation}
where $\mathcal{M}_{rf}^i$ and $\mathcal{M}_{rb}^i$ are foreground and background masks, which present the semantic response between prototype tokens and RGB features. $\odot$ is the dot product operation. Similarity, we can use the same operation to generate foreground mask $\mathcal{M}_{df}^i$ and background mask $\mathcal{M}_{db}^i$ on the depth branch.

To achieve cross-modal information interaction, we group foreground or background masks from the RGB and Depth branches to refine foreground or background regions via a masked linear attention mechanism (MLA). As shown in Fig. 3, the MDA module adopts foreground MLA and background MLA to cooperatively enhance foreground and background on both of RGB and Depth branches:
\begin{equation}
	\begin{split}
		\widetilde{f}_r^i &= \mathrm{Proj}_{\mathcal{C}}([V_r^i \cdot \mathrm{Softmax}((\mathcal{M}_{rf}^i K_r^i)^{\mathrm{T}} (\mathcal{M}_{df}^i Q_r^i)), \\ & \quad \,\, V_r^i \cdot \mathrm{Softmax}((\mathcal{M}_{rb}^i K_r^i)^{\mathrm{T}} (\mathcal{M}_{db}^i Q_r^i))]),
	\end{split}
\end{equation}
where $K_r^i$, $Q_r^i$, and $V_r^i$ are query, key, and value matrices of RGB feature $f_r^i$ generated by three different 1$\times$1 convolutions and a concatenation operation $[,]$. $\widetilde{f}_r^i$ is the output RGB features of the MDA module. We apply the same operation on the depth branch to generate refined Depth features $\widetilde{f}_d^i$. We achieve multi-modality interaction by exchanging the foreground and background images between the RGB and depth branches.

\subsection{Depth Adaptive Injection}
How to integrate RGB and depth features to achieve cross-modal complementarity is crucial for the RGB-D segmentation field. In this paper, we provide a novel perspective to explore decision-making dependency via modality-consistency and modality-specific prototypes generated by the SPE module. As shown in Fig. 4 (b), DAI leverages similarity scores between RGB/Depth modality-specific prototypes and the RGB-depth modality-consistency prototype to measure the correlation of RGB and Depth features in the decision-making process.


Firstly, we obtain multi-modality consistency prototypes via integrating RGB and depth modality-consistency prototypes:
\begin{equation}
	\mathcal{P}_{\mathrm{Con}}^i = \frac{1}{2} \times (\mathcal{FP}_{rc}^i + \mathcal{FP}_{dc}^i),
\end{equation}
where $\mathcal{P}_{\mathrm{Con}}^i$ is the multi-modality consistency prototype. Next, we compute similarity scores between modality-specific prototypes and multi-modality consistency prototype on RGB and Depth branches:
\begin{equation}
	\begin{split}
		\mathcal{W}_r^i &= \mathrm{Cos}(\mathcal{FP}_{rs}^i, \mathcal{P}_{\mathrm{Con}}^i), \\ 
		\mathcal{W}_d^i &= \mathrm{Cos}(\mathcal{FP}_{ds}^i, \mathcal{P}_{\mathrm{Con}}^i),
	\end{split}
\end{equation}
where $\mathcal{W}_r^i$ and $\mathcal{W}_d^i$ determine  which features are shared across modalities and which are modality-specific in both RGB and Depth features.

By disentangling RGB and Depth features into modality-consistency and modality-specific components, we can effectively explore the diversity of multi-modality features. Next, we achieve adaptive fusion of RGB and Depth features:
\begin{equation}
	\begin{split}
		\mathcal{F}_{\mathrm{c}}^i &= \mathcal{W}_r^i \times \widetilde{f}_r^i + \mathcal{W}_d^i \times \widetilde{f}_d^i, \\ 
		\mathcal{F}_f^i &= \mathrm{DC}([(1-\mathcal{W}_r^i)\times \widetilde{f}_r^i, (1-\mathcal{W}_d^i) \times \widetilde{f}_d^i, \mathcal{F}_{\mathrm{c}}^i]),
	\end{split}
\end{equation}
where $\mathcal{F}_{\mathrm{c}}^i$ is modality-consistency features, $(1-\mathcal{W}_r^i)\times \widetilde{f}_r^i$ and $(1-\mathcal{W}_d^i) \times \widetilde{f}_d^i$ are RGB and Depth modality-specific features. $\mathrm{DC}$ and  $\mathcal{F}_f^i$ are a 3$\times$3 deformable convolution and fused features, respectively.

\subsection{Objective}
The total loss $\mathcal{L}_{total}$ of our VCP-DCN contains two components: (1) mask loss between GT and prediction masks from triple decoder; (2) alignment loss in SPE module for learning modality-consistency and modality-specific prototypes, which can be formulated as:
\begin{equation}
	\begin{split}
		\mathcal{L}_{\mathrm{total}} &= \lambda \times \mathcal{L}_{\mathrm{mask}} + \mu \times \mathcal{L}_{\mathrm{align}}, \\ 
		\mathcal{L}_{\mathrm{mask}}  &= \sum_{t}^{\{r,d,f\}} \mathcal{L}_{\mathrm{Hybrid}}(\mathcal{M}_{t}, \mathcal{M}_{\mathrm{GT}}),  \\ 
		\mathcal{L}_{\mathrm{align}} &= \mathcal{L}_{MCL} + \mathcal{L}_{MSL} + \mathcal{L}_{RCL} + \mathcal{L}_{DCL},
	\end{split}
\end{equation}
where $\mathcal{M}_t(t \in \{r,d,f\})$ and $\mathcal{M}_{\mathrm{GT}}$ are prediction mask and GT mask, $\mathcal{L}_{\mathrm{Hybrid}}$ is a hybrid loss with BCE, IoU, and SSIM losses, which are widely used loss functions in image segmentation. $\lambda$ and $\mu$ as weights of $\mathcal{L}_{\mathrm{mask}}$ and $\mathcal{L}_{\mathrm{align}}$ are used to balance the gradient optimization of multi-modality prototype learning and COD-oriented task.

\section{Experiments}

\begin{table*}[t]
	\large
	\renewcommand\arraystretch{2.5}
	\centering
	\fontsize{8}{8}
	\selectfont
	\label{tab:distortion_type}
	\setlength{\tabcolsep}{3pt} 
	\renewcommand\arraystretch{1.0}
	\resizebox*{\textwidth}{!}{
		\large
		\begin{tabular}{lcc|cccc|cccc|cccc}
			\toprule[2pt]
			
			\multirow{2}{*}{\textbf{Method}} &
			\multirow{2}{*}{\textbf{Pub.}} &
			\multirow{2}{*}{\textbf{Tool}} &
			
			\multicolumn{4}{c|}{\textbf{CAMO}} &
			\multicolumn{4}{c|}{\textbf{COD10K}} &
			\multicolumn{4}{c}{\textbf{NC4K}} \cr
			
			& &
			& $S_m\uparrow$ & $E_{\phi}\uparrow$ & $\mathcal{F}_{\beta}^{w}\uparrow$ & $\mathcal{M}\downarrow$ 
			& $S_m\uparrow$ & $E_{\phi}\uparrow$ & $\mathcal{F}_{\beta}^{w}\uparrow$ & $\mathcal{M}\downarrow$ 
			& $S_m\uparrow$ & $E_{\phi}\uparrow$ & $\mathcal{F}_{\beta}^{w}\uparrow$ & $\mathcal{M}\downarrow$ 
			\cr \hline  \hline 
			
			\rowcolor{gray!20} \multicolumn{15}{c}{\textbf{RGB-based COD Methods}} \cr
			
			FSPNet \cite{ref-19} & CVPR23 & DeiT
			& 0.856 & 0.877 & 0.738 & 0.071
			& 0.851 &  0.901 & 0.716 & 0.032
			& 0.879 &  0.913 & 0.789 & 0.044 \cr

			VSCode \cite{ref-21} & CVPR24 & 2D Prompt
			& 0.838 &  0.906 & 0.768 & 0.060
			& 0.847 &  0.913 & 0.744 & 0.028
			& 0.874 &  0.924 & 0.813 & 0.038 \cr

			FSEL \cite{ref-22} & ECCV24 & PVT
			& 0.885 &  0.942  & 0.851 & 0.040
			& 0.873 &  0.928 & 0.800 & 0.021 
			& 0.892 &  0.941 & 0.853 & 0.030  \cr

			CamoFormer \cite{ref-48} & TPAMI24 & PVT
			& 0.872 &  0.931 & 0.831 & 0.046
			& 0.869 &  0.931 & 0.786 & 0.023
			& 0.892 &  0.941 & 0.847 & 0.030  \cr
			
			MCRNet \cite{ref-53} & IJCV25 & VMamba
			& 0.886 &  0.942  & 0.849 & 0.040
			& 0.879 &  0.941  & 0.807 & 0.021
			& 0.893 &  0.943  & 0.853 & 0.030 \cr
			
			CamoDiffusion \cite{ref-54} & TPAMI25 & DM
			& 0.878 &  0.943  & 0.853 & 0.042
			& 0.881 &  0.946 & 0.814 & 0.020
			& 0.893 &  0.942 & 0.859 & 0.029 \cr
			
			CGD \cite{ref-51} & MM25 & CLIP
			& 0.876 &  0.932 & 0.838 & 0.045
			& 0.872 &  0.935 & 0.793 & 0.023
			& 0.889 &  0.935 & 0.847 & 0.032 \cr

			MM-SAM \cite{ref-56} & ICCV25 & BLIP
			& 0.863 & 0.901 & 0.782 & 0.059
			& 0.867 & 0.907 & 0.808 & 0.023
			& - & - & - & -  \cr
			
			ESCNet\cite{ref-55} &  ICCV25 & PVT
			& 0.871 & 0.934 & 0.843 & 0.044 
			& 0.873 & 0.939 & 0.804 & 0.021
			& 0.892 & 0.941 & 0.859 & 0.028 \cr
			
			Samba \cite{ref-57} & CVPR25 & VMamba
			& 0.883 & 0.938 & 0.846 & 0.042
			& 0.865 & 0.935 & 0.791 & 0.022
			& 0.888 & 0.939 & 0.847 & 0.031 \cr
			
			\hline 
			
			\rowcolor{gray!20} \multicolumn{15}{c}{\textbf{RGB-D-based COD Methods via DPT}} \cr 
			
			PopNet \cite{ref-64} & ICCV23 & -
			& 0.806 & 0.869 & - & 0.073
			& 0.827 & 0.897 & - & 0.031
			& 0.852 & 0.908 & - & 0.043 \cr
			
			DaCOD \cite{ref-65} & MM23 & Swin
			& 0.855 & 0.911 & 0.796 & 0.051
			& 0.840 & 0.908 & 0.729 & 0.028 
			& 0.874 & 0.923 & 0.814 & 0.035 \cr
			
			RISNet \cite{ref-3} & CVPR24 & -
			& 0.870 &  0.922 & 0.827 & 0.050
			& 0.873 &  0.931 & 0.779 & 0.025
			& 0.882 &  0.925 & 0.834 & 0.037 \cr
			
			DSAM \cite{ref-38} & MM24 & SAM
			& 0.832 & 0.913 & 0.821 & 0.061 
			& 0.846 & 0.921 & 0.789 & 0.033
			& 0.871 & 0.932 & 0.826 & 0.040 \cr
			
			SAM-COD \cite{ref-50} & ICCV25 & SAM
			& 0.866 & \textbf{0.943} & 0.839 & 0.047
			& \textbf{0.881} & 0.936 & \textbf{0.817} & 0.023
			& 0.889 & 0.935 & 0.847 & 0.032 \cr
			
			Samba \cite{ref-57} & CVPR25 & VMamba
			& 0.871 & 0.932 & 0.841 & 0.045
			& 0.853 & 0.931 & 0.793 & 0.024
			& 0.881 & 0.934 & 0.842 & 0.033 \cr
			
			\textbf{VCP-DCN} & \textbf{Ours} & VMamba
			& \textbf{0.875} & 0.937 & \textbf{0.851} & \textbf{0.041}
			& 0.862 & \textbf{0.938} & 0.799 & \textbf{0.021}
			& \textbf{0.891} & \textbf{0.940} & \textbf{0.849} & \textbf{0.031} \cr

			\rowcolor{gray!20} \multicolumn{15}{c}{\textbf{RGB-D-based COD Methods via Depth Anything V2}} \cr 
			Samba \cite{ref-57} & CVPR25 & VMamba
			& 0.883 & 0.941 & 0.863 & 0.036
			& 0.870 & 0.939 & 0.842 & 0.019
			& 0.887 & 0.939 & 0.863 & 0.027 \cr
			
			\textbf{VCP-DCN} & \textbf{Ours} & VMamba
			& 0.892 & 0.946 & \textbf{0.871} & \textbf{0.033}
			& \textbf{0.881} & \textbf{0.948} & 0.930 & \textbf{0.017}
			& 0.896 & \textbf{0.945} & \textbf{0.871} & \textbf{0.026} \cr
			

			DaCOD \cite{ref-65} & MM23 & Swin
			& 0.877 & 0.941 & 0.846 & 0.042
			& 0.874 & 0.922 & 0.795 & 0.021 
			& 0.883 & 0.938 & 0.847 & 0.030 \cr
			
			\textbf{VCP-DCN} & \textbf{Ours} & Swin
			& \textbf{0.899} & \textbf{0.951} & \textbf{0.871} & \textbf{0.033}
			& 0.880 & 0.947 & \textbf{0.932} & \textbf{0.017}
			& \textbf{0.898} & 0.944 & 0.869 & 0.028 \cr
			
			\bottomrule[2pt]
			
	\end{tabular}}
	\caption{\label{comparison} Quantitative comparison of VCP-DCN with SOTA COD methods on CAMO, COD10K, and NC4K datasets. The best results are highlighted in \textbf{bold} in each setting.} 
\end{table*}

\subsection{Experiment Settings}
\textbf{Datasets and Evaluation Metrics}. Following previous works \cite{ref-19, ref-22, ref-48}, we use three benchmark datasets to evaluate our VCP-DCN, including CAMO \cite{ref-60}, COD10K \cite{ref-61}, and NC4K \cite{ref-62}. Specifically, we select 1000 samples from CAMO and 3040 samples from COD10K as training sets for training models; the rest of CAMO, COD10K, and NC4K are used to evaluate performance. For a fair comparison, we use universal evaluation metrics to measure the performance of different approaches, including Structure-measure ($S_m\uparrow$), E-measure ($E_{\phi}\uparrow$), weighted F-measure ($\mathcal{F}_{\beta}^{w}\uparrow$), and mean absolute error ($\mathcal{M}\downarrow$).

\textbf{Implementation Details}. We follow most of the visual models to adopt a pre-training backbone on ImageNet as the feature encoder. In this work, we use the VMamba-S \cite{ref-25} and Swin-S \cite{ref-6} as the backbone to demonstrate the effect of our VCP-DCN. All the experiments are implemented on an RTX 4090 GPU with 24G memory for training 80 epochs. AdamW optimizer is employed with a weight decay of 0.1 and a $5e^{-5}$ original learning rate, which decreased to one-tenth after 40 epochs. All training and testing images are resized to 384 $\times$ 384, and the batch sizes are 4. The weight factors $\lambda$ and $\mu$ are set to 0.3 and 0.1 in our experiments. Different from previous works \cite{ref-64, ref-50, ref-3, ref-38}, which use DPT \cite{ref-63} to generate depth maps, we introduce Depth Anything V2 \cite{ref-13} to generate high-quality depth maps.

\begin{figure}[t]
	\centering
	\includegraphics[width=0.78\textwidth,height=6cm]{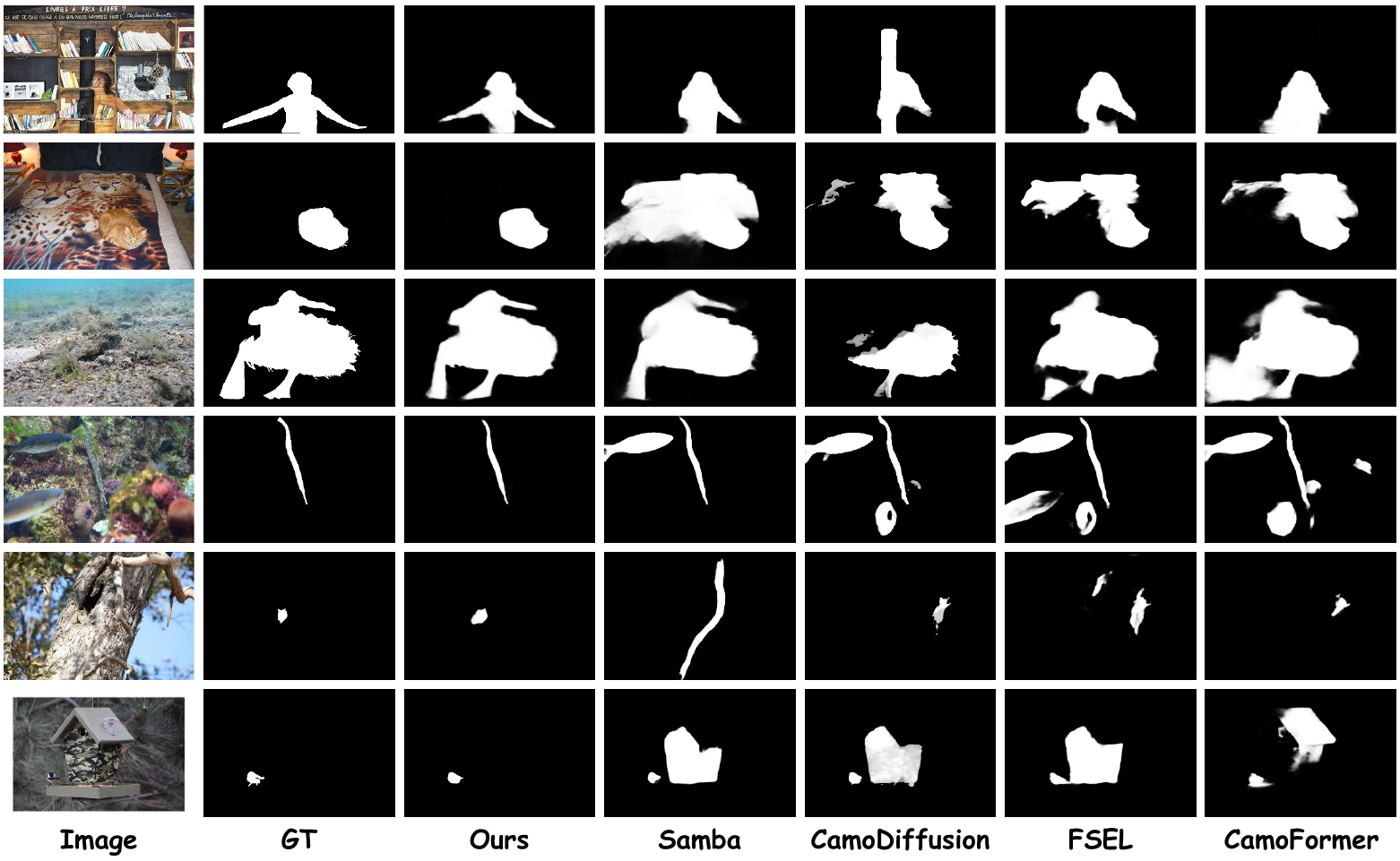}
	\caption{Visual comparisons between our VCP-DCN and several previous SOTA methods, including Samba \cite{ref-57}, CamoDiffusion \cite{ref-54}, FSEL \cite{ref-22}, and CamoFormer \cite{ref-48} on challenging scenes.}
	\label{Fig.3}	
\end{figure}

\begin{table}[t]
	\large
	\renewcommand\arraystretch{1.5}
	\centering
	\fontsize{8}{8}
	\selectfont
	\label{tab:distortion_type}
	\setlength{\tabcolsep}{20pt} 
	\renewcommand\arraystretch{1.0}
	\resizebox{\linewidth}{!}{
		\large
		\begin{tabular}{l|cc|cc|cc}
			\toprule[2pt]
			
			\multirow{2}{*}{\textbf{Method}} &
			\multicolumn{2}{c|}{\textbf{NJU2K}} &
			\multicolumn{2}{c|}{\textbf{NLPR}} &
			\multicolumn{2}{c}{\textbf{DUT}} \cr

			&$\mathcal{F}_{\beta}^{w}\uparrow$ & $\mathcal{M}\downarrow$ 
			&$\mathcal{F}_{\beta}^{w}\uparrow$ & $\mathcal{M}\downarrow$ 
			&$\mathcal{F}_{\beta}^{w}\uparrow$ & $\mathcal{M}\downarrow$ 
			\cr \hline  \hline 
			
			AFB \cite{ref-70}
			& 0.910 & 0.028 & 0.902 & 0.021 & 0.919 & 0.027 \cr
			
			VSCode \cite{ref-21}
			& 0.925 & 0.025 & 0.909 & 0.020 & 0.945 & 0.019 \cr
			
			DIMSOD \cite{ref-68}
			& - & 0.028 & - & - & - & 0.023 \cr
			
			LEAF \cite{ref-67}
			& - & 0.025 & - & 0.016 & - & 0.019 \cr
			
			ACINet \cite{ref-72}
			& 0.923 & 0.025 & 0.919 & 0.016 & 0.944 & 0.021 \cr
			
			TwinsTNet \cite{ref-69}
			& 0.925 & 0.026 & 0.922 & 0.016 & 0.946 & 0.018 \cr
			
			UniSOD \cite{ref-71}
			& 0.922 & 0.025 & 0.909 & 0.019 & 0.952 & 0.017 \cr
			
			VST++ \cite{ref-66}
			& - & - & 0.898 & 0.021
			& 0.929 & 0.025 \cr
			
			VCP-DCN
			& \textbf{0.929} & \textbf{0.023} & \textbf{0.925} & \textbf{0.015} 
			& \textbf{0.957} & \textbf{0.016}  \cr

			\bottomrule[2pt]
			
	\end{tabular}}
	\caption{\label{comparison} Comparison with SOTA RGB-D SOD methods on NJU2K \cite{ref-73}, NLPR \cite{ref-74}, and DUT \cite{ref-75} datasets. The best results are highlighted in \textbf{bold}.}
\end{table}

\subsection{Comparison with State-of-the-Arts}
We conduct a comprehensive performance comparison with COD approaches, which are classified into three groups: (1) RGB-based methods, including FSPNet \cite{ref-19}, VSCode \cite{ref-21}, FSEL \cite{ref-22}, CamoFormer \cite{ref-48}, MCRNet \cite{ref-53}, CamoDiffusion \cite{ref-54}, ESCNet \cite{ref-55} , and Samba \cite{ref-57}; (2) RGB-D-based methods, including PopNet \cite{ref-64}, DaCOD \cite{ref-65} , and RISNet \cite{ref-3}; (3) VFM-based methods with extra modality, including DSAM \cite{ref-38}, CGD \cite{ref-51} , SAM-COD \cite{ref-50}, and MM-SAM \cite{ref-56}.

\textbf{Quantitative Evaluation.} The results of the performance comparison are reported in Table 1.  We report the performance of our VCP-DCN on groups (2) and (3) with pseudo depth maps generated from DPT and depth anything v2, respectively. Firstly, our VCP-DCN significantly outperforms other RGB-D methods, like the latest Samba \cite{ref-57} with the same backbone-VMamba. Furthermore, compared with the VFM-based method SAM-COD \cite{ref-50} with SAM, our VCP-DCN still gets overall advantageous performances with less parameters and FLOPs. Besides, compared with text-based COD methods, like CGD \cite{ref-51} and MM-SAM \cite{ref-56}, our method can keep the performance superiority. Finally, performance comparison of our VCP-DCN on groups (1) and (2) indicates that the quality of depth maps generated by DAMv2 \cite{ref-13} is better than that generated by DPT \cite{ref-63}. \textit{Note that our VCP-DCN consistently outperforms other methods with the same backbone and pseudo depth maps, like Samba and DaCOD.}



\begin{figure}[t]
	\centering
	\includegraphics[width=0.98\textwidth,height=5.5cm]{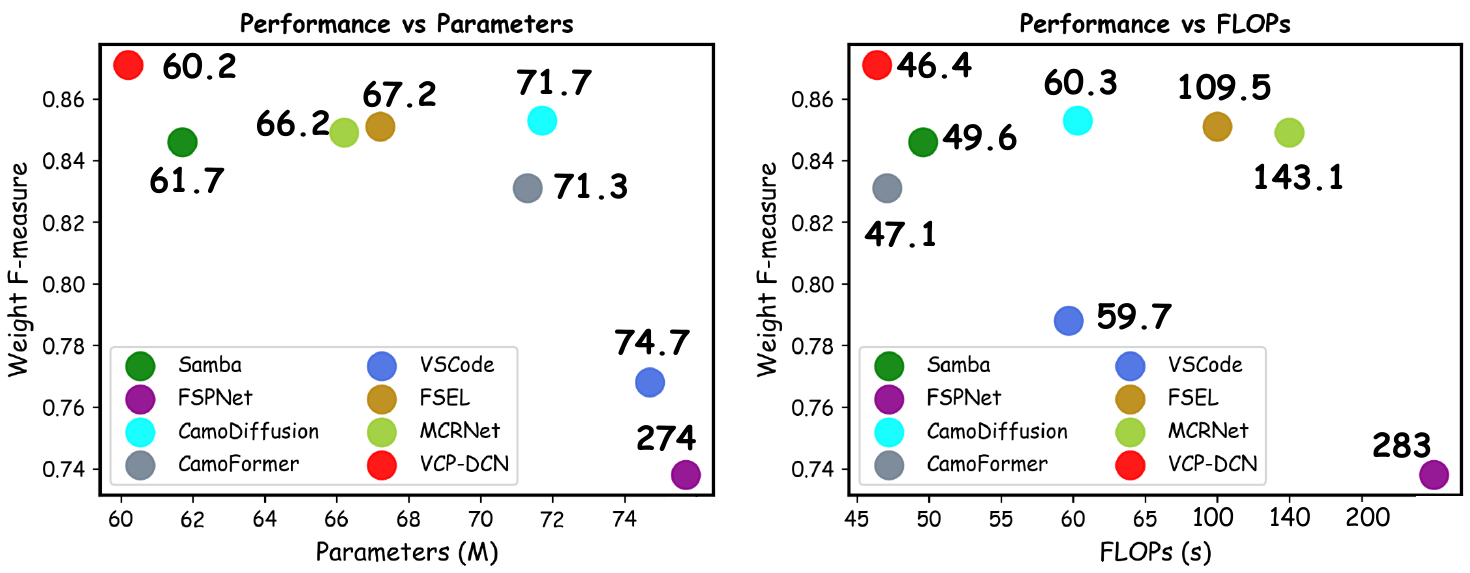}
	\caption{Efficiency comparisons between our VCP-DCN and several SOTA methods on CAMO dataset.}
	\label{Fig.6}	
\end{figure}

\textbf{Qualitative Evaluation.}  As shown in Fig. 5,  we present a visualization comparison of different SOTA COD methods, including Samba \cite{ref-57}, CamoDiffusion \cite{ref-54}, FSEL \cite{ref-22}, and CamoFormer \cite{ref-48}, on several challenging scenes, \textit{e.g.}, artistic disguise, underwater objects, and small objects. These comparisons prove the effectiveness of our VCP-DCN. For example, our method accurately locates and segments the camouflaged cat with a carpet with cat patterns on the 2nd line of Fig. 5, but other methods locate the cat patterns as a camouflaged object.

\textbf{Efficiency Evaluation.} To evaluate the computation efficiency, we make a comprehensive efficiency comparison. As shown in Fig. 6, we report the parameters and FLOPs values of our VCP-DCN and compared methods. The comparison results indicate that our VCP-DCN gets the best performance with the smallest parameters 60.2 M and FLOPs 46.4 G, which further proves the superiority of our method.

\textbf{Application with RGB-D SOD Task.}
We further apply our VCP-DCN to RGB-D Salient Object Detection (SOD). We use a universal training set in the RGB-D SOD field to train our model, including NJU2K \cite{ref-73}, NLPR \cite{ref-74}, and DUT \cite{ref-75}; the rest are used for evaluation. The comparison results are reported in Table 2. Significantly, our VCP-DCN outperforms the compared RGB-D SOD methods, which proves that the depth collaborative network of VCP-DCN is effective for RGB-D segmentation tasks.

\begin{table*}[t]
	\centering
	
	\begin{minipage}[t]{0.49\linewidth}
		\centering
		\fontsize{8}{8}\selectfont
		\setlength{\tabcolsep}{1pt} 
		\renewcommand\arraystretch{1.0}
		
		\resizebox{\linewidth}{!}{
			\begin{tabular}{lccccccc}
				\toprule[2pt]
				
				\multirow{2}{*}{\textbf{Item}} &
				\multirow{2}{*}{\textbf{SPE}} &
				\multirow{2}{*}{\textbf{MDA}} &
				\multirow{2}{*}{\textbf{DAI}} &
				\multicolumn{4}{c}{\textbf{CAMO}} \\
				
				& & & 
				& $S_m$ & $E_{\phi}$ & $\mathcal{F}_{\beta}^{w}\uparrow$ & $\mathcal{M}\downarrow$
				\\ \hline \hline 
				
				C1 &  &  &
				& 0.856 & 0.907 & 0.793 & 0.052  \\
				
				C2 & \checkmark &  &
				& 0.871 & 0.925 & 0.822 & 0.045 \\
				
				C3 & \checkmark &  \checkmark &
				& 0.883 & 0.931 & 0.854 & 0.039 \\
				
				C4 & \checkmark &   & \checkmark
				& 0.881 & 0.934 & 0.841 & 0.041 \\
				
				C5 & \checkmark & \checkmark & \checkmark
				& \textbf{0.892} & \textbf{0.946} & \textbf{0.871} & \textbf{0.033} \\
				
				\bottomrule[2pt]
			\end{tabular}
		}
		
		\caption{Ablation studies of components in our VCP-DCN.}
	\end{minipage}
	\hfill
	\begin{minipage}[t]{0.49\linewidth}
		\centering
		\fontsize{8}{8}\selectfont
		\setlength{\tabcolsep}{1pt} 
		\renewcommand\arraystretch{1.1}
		
		\resizebox{\linewidth}{!}{
			\begin{tabular}{lcccc}
				\toprule[2pt]
				
				\multirow{2}{*}{\textbf{Item}} &
				\multicolumn{4}{c}{\textbf{CAMO}} \\
				
				& $S_m$ & $E_{\phi}$ & $\mathcal{F}_{\beta}^{w}\uparrow$ & $\mathcal{M}\downarrow$
				\\ \hline \hline 
				
				A1: w/o MDA
				& 0.881 & 0.934 & 0.841 & 0.041 \\
				
				A2: A1 w/ Foreground MLA
				& 0.889 & 0.941 & 0.862 & 0.037 \\
				
				A3: A1 w/ Background MLA
				& 0.885 & 0.938 & 0.851 & 0.039 \\
				
				\hline 
				
				B1: w/o DAI
				& 0.883 & 0.931 & 0.854 & 0.039  \\
				
				B2: B1 w/ Consistency MMF
				& 0.887 & 0.938 & 0.861 & 0.038  \\
				
				B3: B1 w/ Specific MMF
				& 0.884 & 0.945 & 0.869 & 0.035  \\
				
				\rowcolor{gray!20} \textit{Ours}
				& \textbf{0.892} & \textbf{0.946} & \textbf{0.871} & \textbf{0.033} \\
				
				\bottomrule[2pt]
			\end{tabular}
		}
		
		\caption{Ablation studies of MDA and DAI. MMF is multi-modality features.}
	\end{minipage}
\end{table*}

%
%
%
%
%
%
%
%
%

\subsection{Ablation Study}
To analyze the contribution of different components in our VCP-DCN, we conduct comprehensive ablation studies on CAMO datasets with $S_m$, $E_{\phi}$, $\mathcal{F}_{\beta}^{w}$, and $\mathcal{M}$ metrics.

\subsubsection{Effectiveness of VCP-DCN.}
To investigate the role of the depth collaborative network in our \( \mathcal{VCP\text{-}DCN} \), we conduct faithful ablation studies in Table 3, aiming to highlight how our method learns multimodal features. First, adding the SPE module to C1 yields significant gains through multi-modality alignment, validating the effectiveness of SPE. Building on this, the modality-specific and modality-consistency prototype tokens generated from the SPE module enable MDA and DAI to obtain further performance improvements (\textit{e.g.}, C3 vs C1: 0.039 vs 0.052 and C4 vs C1: 0.041 vs 0.052). Combining MDA and DAI leads to the best performance. Since SPE underpins both MDA and DAI, we do not discuss the cases of C1 + MDA and C1 + DAI separately.

%
%
%
%
%
%
%
%
%
%
%
%
%
%
%
%
%

\subsubsection{Effectiveness of MDA Module.}
The DMA module uses foreground and background prototype tokens to generate foreground and background masks, which are then used to build foreground- and background-masked linear attention. To validate the effectiveness of dual masked linear attention (F-MLA and B-MLA), we conduct two ablation studies in Table 4. F-MLA and B-MLA boots A1 by 9.8\% and 4.9\% $\mathcal{M}$ scores on CAMO, respectively, prompting accurate localization of camouflaged objects in a complex background.

\subsubsection{Effectiveness of DAI Module.}

DAI module leverages the cosine similarity between modality-specific and modality-consistency prototypes to learn modality-specific RGB, modality-specific depth, and modality-consistency RGB-D features. Ablation studies on Table 4 show that modality-specific and modality-consistency features are valuable for improving multi-modality feature representation. For examples, modality-specific features boots B1 by 12.8\% $\mathcal{M}$ scores on CAMO.

\begin{figure}[t]
	\centering
	\includegraphics[width=0.98\textwidth,height=6cm]{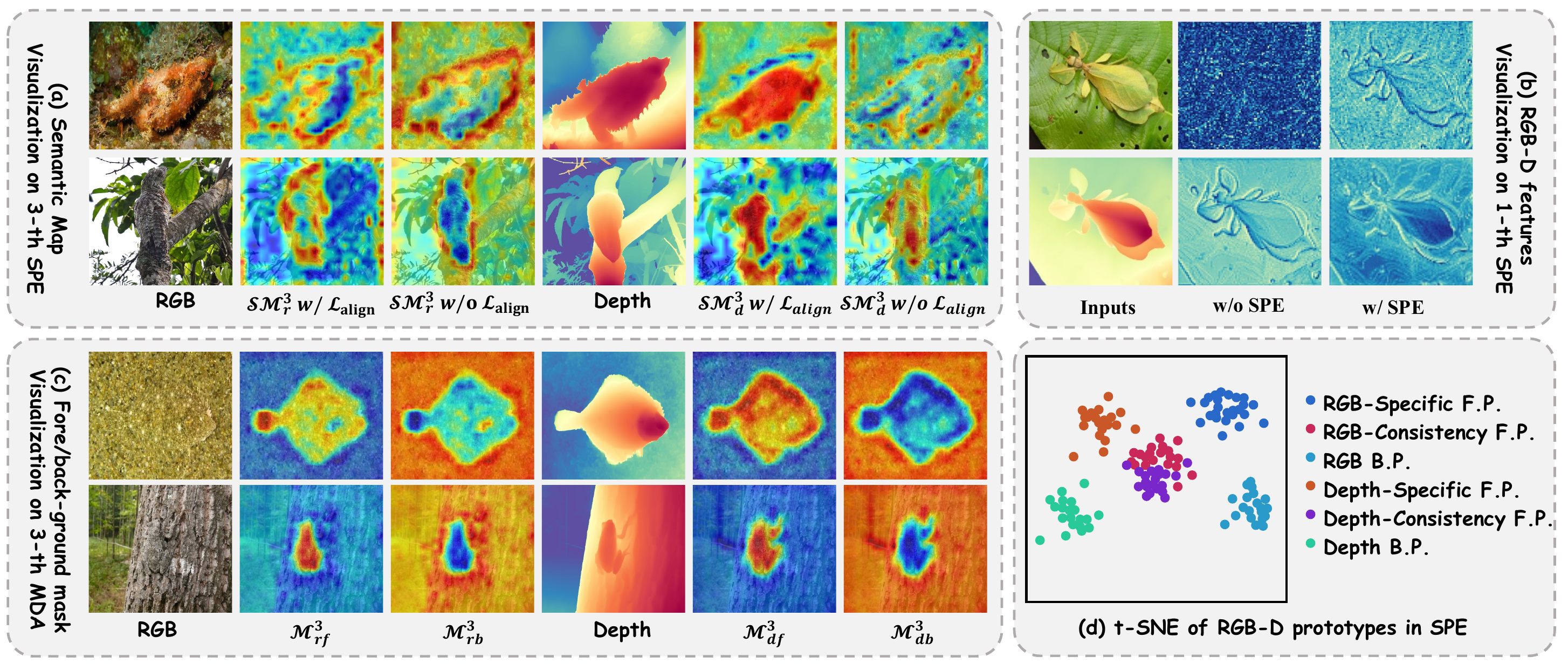}
	\caption{Heatmap visualization of multi-modality semantic maps $\mathcal{SM}_{r}^{3}$ and $\mathcal{SM}_{d}^{3}$ in SPE module.}
	\label{Fig.7}	
\end{figure}


\subsection{Heatmap Visualization}
To provide more comprehensive insights into our VCP-DCN, we visualize the heatmap of multi-modality features around it. As shown in Fig. 7 (a), we visualize the semantic maps $\mathcal{SM}_{r}^3$ and $\mathcal{SM}_{r}^3$ with or without align loss $\mathcal{L}_{\mathrm{align}}$. semantic maps with $\mathcal{L}_{\mathrm{align}}$ obviously outperforms the ones without $\mathcal{L}_{\mathrm{align}}$, which proves that $\mathcal{L}_{\mathrm{align}}$ is effective for improving multi-modality representation. As shown in Fig.7 (b), depth features provides more sharp edges and more detailed cues than RGB features.  SPE uses modality-consistency loss to enhance the edge details of RGB features (w/ SPE \textit{vs} w/o SPE) via reducing the distance between $\mathcal{FP}_{rc}$ and $\mathcal{FP}_{dc}$. Fig. 7 (c) presents foreground and background masks of RGB and Depth features, which provide clear foreground and background regions for enhancing interaction between RGB and Depth features. In Fig. 7 (d), we present t-SNE of RGB-D prototypes, where modality-specific and consistency prototypes are well-separated.

\section{Conclusion}
In this paper, we presents VCP-DCN, a novel COD model, which mines salient properties of camouflaged objects from a depth domain perspective. The core of VCP-DCN is a depth collaborative network, which adopt a progressive strategy to learn multi-modality representation via alignment, interaction, and fusion. The proposed SPE module leverages modality-consistency and modality-specific prototypes to align RGB and depth features. MDA module leverages modality-consistency prototypes to achieve multi-modality interaction. The DAI module uses similarity scores between modality-consistency and modality-specific prototypes to adaptively fuse multi-modality features. Experiments on benchmark datasets demonstrate that our VCP-DCN outperforms the existing 15 SOTA approaches. Furthermore, extra experiments on the RGB-D SOD datasets proves the flexibility of our VCP-DCN on other RGB-D segmentation task.

\section*{Acknowledgements}
This work was supported in part by the National Natural Science Foundation of China under Grant 62372348, in part by the Key Research and Development Program of Shaanxi under Grant 2024GXZDCYL-02-10, in part by Scientific and Technological Innovation Teams in Shaanxi Province under Grant 2025RS-CXTD-011, in part by the Shaanxi Province Core Technology Research and Development Project under Grant 2024QY2-GJHX-11, in part by the Establishment Fund of the State Key Laboratory of Wakefullness/Sleep and Cognition(Chongqing Institute for Brain and Intelligence), in part by the Fundamental Research Funds for the Central Universities under Grant QTZX26138.

%
%
\bibliographystyle{splncs04}
\bibliography{main}

\begin{thebibliography}{10}
\providecommand{\url}[1]{\texttt{#1}}
\providecommand{\urlprefix}{URL }
\providecommand{\doi}[1]{https://doi.org/#1}

\bibitem{ref-43}
Chen, G., Wang, Q., Dong, B., Ma, R., Liu, N., Fu, H., Xia, Y.: Em-trans:
  Edge-aware multimodal transformer for rgb-d salient object detection. IEEE
  Transactions on Neural Networks and Learning Systems  \textbf{36}(2),
  3175--3188 (2024)

\bibitem{ref-47}
Cheng, H., Luo, J., Zhang, X.: Multimodal industrial anomaly detection via
  uni-modal and cross-modal fusion. IEEE Transactions on Industrial Informatics
   (2025)

\bibitem{ref-42}
Cong, R., Liu, H., Zhang, C., Zhang, W., Zheng, F., Song, R., Kwong, S.:
  Point-aware interaction and cnn-induced refinement network for rgb-d salient
  object detection. In: Proceedings of the ACM International Conference on
  Multimedia. pp. 406--416 (2023)

\bibitem{ref-20}
Cong, R., Sun, M., Zhang, S., Zhou, X., Zhang, W., Zhao, Y.: Frequency
  perception network for camouflaged object detection. In: Proceedings of the
  ACM International Conference on Multimedia. pp. 1179--1189 (2023)

\bibitem{ref-1}
Cuthill, I.C., Stevens, M., Sheppard, J., Maddocks, T., P{\'a}rraga, C.A.,
  Troscianko, T.S.: Disruptive coloration and background pattern matching.
  Nature  \textbf{434}(7029),  72--74 (2005)

\bibitem{ref-4}
Dong, B., Wang, W., Fan, D.P., Li, J., Fu, H., Shao, L.: Polyp-pvt: Polyp
  segmentation with pyramid vision transformers. arXiv preprint
  arXiv:2108.06932  (2021)

\bibitem{ref-32}
Duan, S., Yang, X., Wang, N., Gao, X.: Lightweight rgb-d salient object
  detection from a speed-accuracy tradeoff perspective. IEEE Transactions on
  Image Processing  (2025)

\bibitem{ref-59}
Duan, S., Yang, X., Wang, N., Gao, X.: Lightweight rgb-d salient object
  detection from a speed-accuracy tradeoff perspective. IEEE Transactions on
  Image Processing  (2025)

\bibitem{ref-15}
Fan, D.P., Ji, G.P., Cheng, M.M., Shao, L.: Concealed object detection. IEEE
  Transactions on Pattern Analysis and Machine Intelligence  \textbf{44}(10),
  6024--6042 (2021)

\bibitem{ref-14}
Fan, D.P., Ji, G.P., Sun, G., Cheng, M.M., Shen, J., Shao, L.: Camouflaged
  object detection. In: Proceedings of the IEEE/CVF Conference on Computer
  Vision and Pattern Recognition. pp. 2777--2787 (2020)

\bibitem{ref-61}
Fan, D.P., Ji, G.P., Sun, G., Cheng, M.M., Shen, J., Shao, L.: Camouflaged
  object detection. In: Proceedings of the IEEE/CVF Conference on Computer
  Vision and Pattern Recognition. pp. 2777--2787 (2020)

\bibitem{ref-30}
Hao, Z., Xiao, Z., Luo, Y., Guo, J., Wang, J., Shen, L., Hu, H.: Primkd:
  Primary modality guided multimodal fusion for rgb-d semantic segmentation.
  In: Proceedings of the ACM International Conference on Multimedia. pp.
  1943--1951 (2024)

\bibitem{ref-28}
He, J., Fu, K., Liu, X., Zhao, Q.: Samba: A unified mamba-based framework for
  general salient object detection. In: Proceedings of the Computer Vision and
  Pattern Recognition Conference. pp. 25314--25324 (2025)

\bibitem{ref-57}
He, J., Fu, K., Liu, X., Zhao, Q.: Samba: A unified mamba-based framework for
  general salient object detection. In: Proceedings of the Computer Vision and
  Pattern Recognition Conference. pp. 25314--25324 (June 2025)

\bibitem{ref-5}
He, K., Zhang, X., Ren, S., Sun, J.: Deep residual learning for image
  recognition. In: Proceedings of the IEEE/CVF Conference on Computer Vision
  and Pattern Recognition. pp. 770--778 (2016)

\bibitem{ref-19}
Huang, Z., Dai, H., Xiang, T.Z., Wang, S., Chen, H.X., Qin, J., Xiong, H.:
  Feature shrinkage pyramid for camouflaged object detection with transformers.
  In: Proceedings of the IEEE/CVF Conference on Computer Vision and Pattern
  Recognition. pp. 5557--5566 (2023)

\bibitem{ref-75}
Ji, W., Yan, G., Li, J., Piao, Y., Yao, S., Zhang, M., Cheng, L., Lu, H.: Dmra:
  Depth-induced multi-scale recurrent attention network for rgb-d saliency
  detection. IEEE Transactions on Image Processing  \textbf{31},  2321--2336
  (2022)

\bibitem{ref-73}
Ju, R., Ge, L., Geng, W., Ren, T., Wu, G.: Depth saliency based on anisotropic
  center-surround difference. In: IEEE International Conference on Image
  Processing. pp. 1115--1119. IEEE (2014)

\bibitem{ref-37}
Kirillov, A., Mintun, E., Ravi, N., Mao, H., Rolland, C., Gustafson, L., Xiao,
  T., Whitehead, S., Berg, A.C., Lo, W.Y., et~al.: Segment anything. In:
  Proceedings of the IEEE/CVF International Conference on Computer Vision. pp.
  4015--4026 (2023)

\bibitem{ref-60}
Le, T.N., Nguyen, T.V., Nie, Z., Tran, M.T., Sugimoto, A.: Anabranch network
  for camouflaged object segmentation. Computer Vision and Image Understanding
  \textbf{184},  45--56 (2019)

\bibitem{ref-44}
Li, Y., Zhu, Z., Zhang, Y., Chen, Y., Yu, Z.: Boost the inference with
  co-training: A depth-guided mutual learning framework for semi-supervised
  medical polyp segmentation. In: Proceedings of the IEEE/CVF Conference on
  Computer Vision and Pattern Recognition. pp. 10394--10403 (2025)

\bibitem{ref-2}
Lian, S., Zhang, Z., Li, H., Li, W., Yang, L.T., Kwong, S., Cong, R.: Diving
  into underwater: Segment anything model guided underwater salient instance
  segmentation and a large-scale dataset. In: Proceedings of the International
  Conference on Machine Learning. pp. 29545--29559 (2024)

\bibitem{ref-50}
Liu, J., Kong, L., Chen, G.: Improving sam for camouflaged object detection via
  dual stream adapters. In: Proceedings of the IEEE/CVF International
  Conference on Computer Vision (2025)

\bibitem{ref-36}
Liu, J., Kong, L., Chen, G.: Improving sam for camouflaged object detection via
  dual stream adapters. In: Proceedings of the IEEE/CVF International
  Conference on Computer Vision (2025)

\bibitem{ref-66}
Liu, N., Luo, Z., Zhang, N., Han, J.: Vst++: Efficient and stronger visual
  saliency transformer. IEEE Transactions on Pattern Analysis and Machine
  Intelligence  \textbf{46}(11),  7300--7316 (2025)

\bibitem{ref-25}
Liu, Y., Tian, Y., Zhao, Y., Yu, H., Xie, L., Wang, Y., Ye, Q., Jiao, J., Liu,
  Y.: Vmamba: Visual state space model. Proceedings of the Advances in Neural
  Information Processing Systems  \textbf{37},  103031--103063 (2024)

\bibitem{ref-6}
Liu, Z., Lin, Y., Cao, Y., Hu, H., Wei, Y., Zhang, Z., Lin, S., Guo, B.: Swin
  transformer: Hierarchical vision transformer using shifted windows. In:
  Proceedings of the IEEE/CVF International Conference on Computer Vision. pp.
  10012--10022 (2021)

\bibitem{ref-7}
Liu, Z., Mao, H., Wu, C.Y., Feichtenhofer, C., Darrell, T., Xie, S.: A convnet
  for the 2020s. In: Proceedings of the IEEE/CVF Conference on Computer Vision
  and Pattern Recognition. pp. 11976--11986 (2022)

\bibitem{ref-21}
Luo, Z., Liu, N., Zhao, W., Yang, X., Zhang, D., Fan, D.P., Khan, F., Han, J.:
  Vscode: General visual salient and camouflaged object detection with 2d
  prompt learning. In: Proceedings of the IEEE/CVF Conference on Computer
  Vision and Pattern Recognition. pp. 17169--17180 (2024)

\bibitem{ref-62}
Lv, Y., Zhang, J., Dai, Y., Li, A., Liu, B., Barnes, N., Fan, D.P.:
  Simultaneously localize, segment and rank the camouflaged objects. In:
  Proceedings of the IEEE/CVF Conference on Computer Vision and Pattern
  Recognition. pp. 11591--11601 (2021)

\bibitem{ref-69}
Lyu, P., Yu, X., Chi, J., Wu, H., Wu, C., Rajapakse, J.C.: Twinstnet:
  Broad-view twins transformer network for bi-modal salient object detection.
  IEEE Transactions on Image Processing  (2025)

\bibitem{ref-46}
Mao, K., Wei, P., Lian, Y., Wang, Y., Zheng, N.: Beyond single-modal boundary:
  Cross-modal anomaly detection through visual prototype and harmonization. In:
  Proceedings of the IEEE/CVF Conference on Computer Vision and Pattern
  Recognition. pp. 9964--9973 (2025)

\bibitem{ref-17}
Mei, H., Ji, G.P., Wei, Z., Yang, X., Wei, X., Fan, D.P.: Camouflaged object
  segmentation with distraction mining. In: Proceedings of the IEEE/CVF
  Conference on Computer Vision and Pattern Recognition. pp. 8772--8781 (2021)

\bibitem{ref-11}
Oquab, M., Darcet, T., Moutakanni, T., Vo, H., Szafraniec, M., Khalidov, V.,
  Fernandez, P., Haziza, D., Massa, F., El-Nouby, A., et~al.: Dinov2: Learning
  robust visual features without supervision. arXiv preprint arXiv:2304.07193
  (2023)

\bibitem{ref-26}
Pei, X., Huang, T., Xu, C.: Efficientvmamba: Atrous selective scan for light
  weight visual mamba. In: Proceedings of the AAAI Conference on Artificial
  Intelligence. vol.~39, pp. 6443--6451 (2025)

\bibitem{ref-74}
Peng, H., Li, B., Xiong, W., Hu, W., Ji, R.: Rgbd salient object detection: A
  benchmark and algorithms. In: Proceedings of the European Conference on
  Computer Vision. pp. 92--109. Springer (2014)

\bibitem{ref-63}
Ranftl, R., Bochkovskiy, A., Koltun, V.: Vision transformers for dense
  prediction. In: Proceedings of the IEEE/CVF International Conference on
  Computer Vision. pp. 12179--12188 (2021)

\bibitem{ref-56}
Ren, G., Liu, H., Lazarou, M., Stathaki, T.: Multi-modal segment anything model
  for camouflaged scene segmentation. In: Proceedings of the IEEE/CVF
  International Conference on Computer Vision. pp. 19882--19892 (October 2025)

\bibitem{ref-24}
Rombach, R., Blattmann, A., Lorenz, D., Esser, P., Ommer, B.: High-resolution
  image synthesis with latent diffusion models. In: Proceedings of the IEEE/CVF
  Conference on Computer Vision and Pattern Recognition. pp. 10684--10695
  (2022)

\bibitem{ref-72}
Su, Y., Gao, H., Wang, M., Wang, F.: Asymmetric cross-modality interaction
  network for rgb-d salient object detection. Expert Systems with Applications
  \textbf{275},  127004 (2025)

\bibitem{ref-54}
Sun, K., Chen, Z., Lin, X., Sun, X., Liu, H., Ji, R.: Conditional diffusion
  models for camouflaged and salient object detection. IEEE Transactions on
  Pattern Analysis and Machine Intelligence  (2025)

\bibitem{ref-22}
Sun, Y., Xu, C., Yang, J., Xuan, H., Luo, L.: Frequency-spatial entanglement
  learning for camouflaged object detection. In: Proceedings of the European
  Conference on Computer Vision. pp. 343--360 (2024)

\bibitem{ref-71}
Wang, K., Tu, Z., Li, C., Liu, Z., Luo, B.: Unified-modal salient object
  detection via adaptive prompt learning. IEEE Transactions on Circuits and
  Systems for Video Technology  (2025)

\bibitem{ref-70}
Wang, K., Tu, Z., Li, C., Zhang, C., Luo, B.: Learning adaptive fusion bank for
  multi-modal salient object detection. IEEE Transactions on Circuits and
  Systems for Video Technology  \textbf{34}(8),  7344--7358 (2024)

\bibitem{ref-3}
Wang, L., Yang, J., Zhang, Y., Wang, F., Zheng, F.: Depth-aware concealed crop
  detection in dense agricultural scenes. In: Proceedings of the IEEE/CVF
  Conference on Computer Vision and Pattern Recognition. pp. 17201--17211
  (2024)

\bibitem{ref-65}
Wang, Q., Yang, J., Yu, X., Wang, F., Chen, P., Zheng, F.: Depth-aided
  camouflaged object detection. In: Proceedings of the ACM International
  Conference on Multimedia. pp. 3297--3306 (2023)

\bibitem{ref-40}
Wang, Q., Yang, J., Yu, X., Wang, F., Chen, P., Zheng, F.: Depth-aided
  camouflaged object detection. In: Proceedings of the ACM international
  Conference on Multimedia. pp. 3297--3306 (2023)

\bibitem{ref-9}
Wang, W., Xie, E., Li, X., Fan, D.P., Song, K., Liang, D., Lu, T., Luo, P.,
  Shao, L.: Pyramid vision transformer: A versatile backbone for dense
  prediction without convolutions. In: Proceedings of the IEEE/CVF
  International Conference on Computer Vision. pp. 568--578 (2021)

\bibitem{ref-67}
Wu, L., Gao, Z., Fei, H., Lee, M.L., Hsu, W.: Leaf-mamba: Local emphatic and
  adaptive fusion state space model for rgb-d salient object detection. In:
  Proceedings of the ACM International Conference on Multimedia (2025)

\bibitem{ref-64}
Wu, Z., Paudel, D.P., Fan, D.P., Wang, J., Wang, S., Demonceaux, C., Timofte,
  R., Van~Gool, L.: Source-free depth for object pop-out. In: Proceedings of
  the IEEE/CVF International Conference on Computer Vision. pp. 1032--1042
  (2023)

\bibitem{ref-39}
Wu, Z., Paudel, D.P., Fan, D.P., Wang, J., Wang, S., Demonceaux, C., Timofte,
  R., Van~Gool, L.: Source-free depth for object pop-out. In: Proceedings of
  the IEEE/CVF International Conference on Computer Vision. pp. 1032--1042
  (2023)

\bibitem{ref-33}
Xia, C., Duan, S., Ge, B., Zhang, H., Li, K.C.: Hdnet: Multi-modality
  hierarchy-aware decision network for rgb-d salient object detection. IEEE
  Signal Processing Letters  \textbf{29},  2577--2581 (2022)

\bibitem{ref-41}
Xia, C., Duan, S., Fang, X., Gao, X., Sun, Y., Ge, B., Zhang, H., Li, K.C.:
  Efgnet: Encoder steered multi-modality feature guidance network for rgb-d
  salient object detection. Digital Signal Processing  \textbf{131},  103775
  (2022)

\bibitem{ref-8}
Yang, J., Li, C., Dai, X., Gao, J.: Focal modulation networks. Proceedings of
  the Advances in Neural Information Processing Systems  \textbf{35},
  4203--4217 (2022)

\bibitem{ref-31}
Yang, J., Bai, L., Sun, Y., Tian, C., Mao, M., Wang, G.: Pixel difference
  convolutional network for rgb-d semantic segmentation. IEEE Transactions on
  Circuits and Systems for Video Technology  \textbf{34}(3),  1481--1492 (2023)

\bibitem{ref-13}
Yang, L., Kang, B., Huang, Z., Zhao, Z., Xu, X., Feng, J., Zhao, H.: Depth
  anything v2. Proceedings of the Advances in Neural Information Processing
  Systems  \textbf{37},  21875--21911 (2024)

\bibitem{ref-48}
Yin, B., Zhang, X., Fan, D.P., Jiao, S., Cheng, M.M., Van~Gool, L., Hou, Q.:
  Camoformer: Masked separable attention for camouflaged object detection. IEEE
  Transactions on Pattern Analysis and Machine Intelligence  (2024)

\bibitem{ref-38}
Yu, Z., Zhang, X., Zhao, L., Bin, Y., Xiao, G.: Exploring deeper! segment
  anything model with depth perception for camouflaged object detection. In:
  Proceedings of the ACM International Conference on Multimedia. pp. 4322--4330
  (2024)

\bibitem{ref-45}
Yuan, H., Li, X., Yang, Y., Cheng, G., Zhang, J., Tong, Y., Zhang, L., Tao, D.:
  Polyphonicformer: Unified query learning for depth-aware video panoptic
  segmentation. In: Proceedings of the European Conference on Computer Vision.
  pp. 582--599 (2022)

\bibitem{ref-16}
Zhai, Q., Li, X., Yang, F., Chen, C., Cheng, H., Fan, D.P.: Mutual graph
  learning for camouflaged object detection. In: Proceedings of the IEEE/CVF
  Conference on Computer Vision and Pattern Recognition. pp. 12997--13007
  (2021)

\bibitem{ref-51}
Zhang, C., Zhang, Q., Wu, J., Pang, Y.: Cgcod: Class-guided camouflaged object
  detection. In: Proceedings of the ACM International Conference on Multimedia
  (2025)

\bibitem{ref-29}
Zhang, D., Cheng, L., Liu, Y., Wang, X., Han, J.: Mamba capsule routing towards
  part-whole relational camouflaged object detection. International Journal of
  Computer Vision pp. 1--21 (2025)

\bibitem{ref-53}
Zhang, D., Cheng, L., Liu, Y., Wang, X., Han, J.: Mamba capsule routing towards
  part-whole relational camouflaged object detection. International Journal of
  Computer Vision pp. 1--21 (2025)

\bibitem{ref-68}
Zhang, S., Huang, J., Tang, W., Wu, Y., Hu, T., Xu, X., Liu, J.: Dimsod: A
  diffusion-based framework for multi-modal salient object detection. In:
  Proceedings of the AAAI Conference on Artificial Intelligence. vol.~39, pp.
  10103--10111 (2025)

\bibitem{ref-23}
Zhao, J., Li, X., Yang, F., Zhai, Q., Luo, A., Jiao, Z., Cheng, H.:
  Focusdiffuser: Perceiving local disparities for camouflaged object detection.
  In: Proceedings of the European Conference on Computer Vision. pp. 181--198
  (2024)

\bibitem{ref-58}
Zhou, T., Fu, H., Chen, G., Zhou, Y., Fan, D.P., Shao, L.:
  Specificity-preserving rgb-d saliency detection. In: Proceedings of the
  IEEE/CVF International Conference on Computer Vision. pp. 4681--4691 (2021)

\bibitem{ref-18}
Zhou, T., Zhou, Y., Gong, C., Yang, J., Zhang, Y.: Feature aggregation and
  propagation network for camouflaged object detection. IEEE Transactions on
  Image Processing  \textbf{31},  7036--7047 (2022)

\bibitem{ref-55}
Zhou, Z., Li, Y., Zhong, C., Huang, J., Pei, J., Li, H., Tang, H.: Rethinking
  detecting salient and camouflaged objects in unconstrained scenes. In:
  Proceedings of the IEEE/CVF International Conference on Computer Vision. pp.
  22372--22382 (October 2025)

\bibitem{ref-27}
Zhu, L., Liao, B., Zhang, Q., Wang, X., Liu, W., Wang, X.: Vision mamba:
  Efficient visual representation learning with bidirectional state space
  model. arXiv preprint arXiv:2401.09417  (2024)

\bibitem{ref-35}
Zhu, X.F., Xu, T., Atito, S., Awais, M., Wu, X.J., Feng, Z., Kittler, J.:
  Self-supervised learning for rgb-d object tracking. Pattern Recognition
  \textbf{155},  110543 (2024)

\bibitem{ref-34}
Zhu, X.F., Xu, T., Tang, Z., Wu, Z., Liu, H., Yang, X., Wu, X.J., Kittler, J.:
  Rgbd1k: A large-scale dataset and benchmark for rgb-d object tracking. In:
  Proceedings of the AAAI Conference on Artificial Intelligence. vol.~37, pp.
  3870--3878 (2023)

\end{thebibliography}
\end{document}